\documentclass[acmsmall,authorversion,nonacm]{acmart}

\makeatletter
\def\@copyrightspace{\relax}
\makeatother

\usepackage{microtype}
\usepackage{graphicx}
\usepackage{caption}
\usepackage{subcaption}
\usepackage{booktabs}
\usepackage{tabularx}
\usepackage{amsmath}

\usepackage{hyperref}
\usepackage[section]{placeins}

\begin{document}

\title[Plenty of Room in the Middle]{Plenty of Room in the Middle: \texorpdfstring{\\}{ }Revealing Opportunities on a GPU Cluster via Fine-Grained Monitoring}

\author{Andrew Sheinberg}
\authornote{Both authors contributed equally to this research.}
\orcid{0009-0007-7575-3348}
\email{asheinb@princeton.edu}
\author{Samyak Gupta}
\authornotemark[1]
\orcid{0000-0002-3258-0789}
\email{samyakg@princeton.edu}
\author{Kai Li}
\orcid{0000-0002-2095-7024}
\email{li@cs.princeton.edu}
\affiliation{
  \institution{Princeton University}
  \department{Department of Computer Science}
  \city{Princeton}
  \state{New Jersey}
  \country{USA}}
\vskip 0.3in



\begin{abstract}

GPU clusters have become essential computing resources for AI and scientific research in academia and industry. However, their high cost and growing demand often result in long queueing delays. While previous monitoring efforts have operated at second- or minute-level frequencies, they primarily informed job-level scheduling for clusters and lacked the detail needed to optimize GPU-level utilization.

This paper presents an in-depth study of a university research GPU cluster, analyzing resource usage at 100~ms intervals. We have developed a lightweight monitoring tool that leverages Nvidia's DCGM library to capture fine-grained ``compute utilization'' metrics within GPUs. Deployed on a cluster with 89 nodes and 319 GPUs, our tool collected data over several months. This paper focuses on a 10-day period with 7,399 jobs submitted by 150 users from 12 departments.
Our study reveals three key findings: (1) the GPU cluster exhibits significant underutilization, with over 70\% of jobs using less than 20\% of GPU memory and 50\% averaging below 5\% Streaming Multiprocessor (SM) utilization; (2) 80\% of jobs have a drastic mismatch between their requested and actual durations, significantly limiting the scheduler's effectiveness; and (3) the multi-GPU jobs we examined show highly periodic communication patterns. These findings suggest there is plenty of room in the middle: substantial utilization gains are available through better job scheduling and GPU multitasking, independent of hardware or application improvements.

Our monitoring solution is available as open-source software on \href{https://github.com/als244/gpu_server_monitoring}{GitHub}.

\end{abstract}

\maketitle

\section{Introduction}
\label{introduction}
GPU clusters have become the engines of modern AI, powering services and research across diverse domains. The insatiable appetite of large language models and the burgeoning demands of AI applications in science and other fields are driving an exponential growth in computing requirements. While GPU vendors continue to deliver increasingly powerful hardware with impressive performance, the associated costs have skyrocketed. For instance, Nvidia's Blackwell GPUs command a hefty price tag of \$30,000--\$40,000~\cite{cnbc_blackwell_price}. However, GPU clusters are significantly underutilized~\cite{hu_characterization_2021, jeon_analysis_2019,xiao_antman_2020}, raising the critical question: How can we substantially improve GPU cluster utilization?

In many batch-oriented GPU clusters, GPUs are allocated exclusively to individual jobs for the duration of their reservations~\cite{xiao_antman_2020}. Under this allocation model, a GPU remains unavailable to other jobs even when its assigned workload uses only a small fraction of its compute or memory resources. Sharing a GPU across independent jobs---through temporal multiplexing, concurrent execution, or a combination of the two---offers an opportunity to recover this unused capacity.

Modern GPUs provide mechanisms for such sharing, but their hardware and runtime support imposes tradeoffs among allocation flexibility, performance isolation, and memory management. For example, NVIDIA's Multi-Process Service (MPS)~\cite{nvidia_mps} enables concurrent execution across processes, while Multi-Instance GPU (MIG)~\cite{nvidia_mig} provides stronger isolation through predefined hardware partitions. These mechanisms provide useful building blocks, but efficiently adapting resource allocations to changing job demands remains a systems challenge.

GPU memory introduces an additional constraint. Although GPUs support virtual addressing and memory protection, efficiently sharing their limited high-bandwidth memory requires managing the combined memory demands of multiple jobs. Conventional device allocations are not automatically pageable to host memory~\cite{nvidia_cuda_runtime_12_4_1}, while Unified Memory supports oversubscription with potentially substantial page-fault and data-migration costs~\cite{nvidia_cuda_programming_12_4_1}.

These challenges motivate a closer examination of when GPU resources are unused, how long those intervals last, and which workloads could share resources effectively.

Previous monitoring studies have relied on coarse-grained metrics, typically with sampling intervals of seconds to minutes, providing limited insights into the fine-grained utilization patterns of GPU resources.
To close this gap,
we have developed a fine-grained monitoring tool capturing utilization data at 100~ms intervals, the finest granularity supported by Nvidia's Data Center GPU Manager (DCGM)~\cite{nvidia_dcgm} and the same length as Linux's default round-robin time slice. We have deployed this lightweight, scalable tool across a university research cluster of 89 nodes and 319 GPUs for several months.

This paper analyzes data collected during a 10-day observation period,\footnote{This study was conducted in fall 2024; the measurements, hardware, and related work discussed reflect that period.} encompassing 7,399 jobs submitted by 150 users across 12 departments. The analysis reveals three key findings:
\begin{itemize}
    \item
{\bf Significant underutilization}: Over 70\% of jobs used less than 20\% of GPU memory, and 50\% averaged below 5\% SM utilization. This underutilization highlights a substantial gap between the capabilities of modern GPUs and the actual resource demands of many workloads, leaving much of the cluster's computational power untapped.
\item
{\bf Pervasive mismatch between requested and actual job durations}:
Approximately 80\% of jobs show a significant mismatch between their requested and actual runtimes, with one-third hitting their requested time limit. This discrepancy results in poor scheduling decisions, including inflated queue times and inefficient resource allocation, fundamentally undermining the effectiveness of current schedulers.

\item
{\bf Highly periodic communication patterns in multi-GPU jobs}: The multi-GPU workloads we examined demonstrated recurring communication patterns, including predictable bursts of NVLink traffic and synchronization phases. These periodic behaviors present opportunities for system designers to develop predictive resource allocation strategies, minimizing interference and optimizing the scheduling of tasks.
\end{itemize}

These findings show great opportunities to improve GPU cluster efficiency. Much work on computing efficiency has focused on the bottom of the stack (faster hardware)~\cite{feynman_room_1960} or the top (better-optimized applications)~\cite{leiserson_room_2020}. Our results show there is also plenty of room in the middle: the systems layer that allocates, schedules, and multiplexes GPUs across jobs can recover much of the capacity that today's clusters leave idle. Using insights from fine-grained monitoring data, system designers can build scheduling and multitasking solutions that improve GPU utilization while preserving performance isolation, making research GPU clusters more effective and cost-efficient.

\section{Monitoring Framework and Environment}
\label{sec:monitoring}

We have developed and deployed a fine-grained monitoring tool to analyze activities on a university GPU cluster.

\subsection{GPU Cluster Overview}

We conducted our study on a university research cluster comprising 89 nodes interconnected through a shared general-purpose networked file system~\cite{gpfs}. The cluster is shared by researchers across multiple departments and includes the following hardware configurations:
\begin{itemize}
    \item High-Performance Nodes (20): Each equipped with AMD EPYC Rome CPUs (128 cores), 768 GB RAM, and two NVIDIA A100 PCIe 40~GB GPUs.
    \item Standard Nodes (69): Each featuring Intel Ice Lake CPUs (48 cores), 100 GB RAM, and four NVIDIA A100 SXM4 80~GB GPUs.
\end{itemize}
All nodes are equipped with a 100 Gb/s ConnectX-6 InfiniBand NIC and a 1 Gb/s Ethernet NIC for out-of-band traffic.

\subsection{Design Goals and Challenges}

To effectively analyze GPU utilization without disrupting cluster operations, we established the following key design goals for the monitoring system:

\begin{itemize}
    \item Minimal Overhead: The monitoring tool must have minimal impact on cluster performance, with low CPU, memory, and storage usage, as data collection was planned over several months across hundreds of GPUs.
    \item Operational Simplicity: The tool needed to be straightforward to deploy and operate, requiring minimal configuration and maintenance to avoid burdening system administrators.
    \item Fine-Grained Metrics: Capturing high-resolution metrics for Streaming Multiprocessors (SMs) utilization, memory utilization, SM occupancy, PCIe traffic, NVLink traffic, and Tensor Core usage, providing a comprehensive view of GPU performance.
    \item Robust and scalable: The tool should be resilient to node failures and deployable to a large-scale GPU cluster.
\end{itemize}

We have observed that
the standard \texttt{nvidia-smi} metric, commonly used to measure GPU utilization, often misrepresents actual resource usage. It reports only the fraction of cycles during which a kernel is running or a copy engine is active, failing to account for fractional SM usage. For example, a job using just one SM out of 108 on an A100 GPU could still be reported as having 100\% utilization.

To address this limitation, we utilized the ``SM Activity'' metric from Nvidia's DCGM API~\cite{nvidia_dcgm}, which provides a more precise measure of GPU compute utilization. This fine-grained metric allowed us to observe intra-GPU spatial utilization and uncover significant discrepancies in reported utilization.

\begin{figure}[!htbp]
        \centering
         \begin{subfigure}[t]{0.48\textwidth}
                \includegraphics[width=\textwidth]{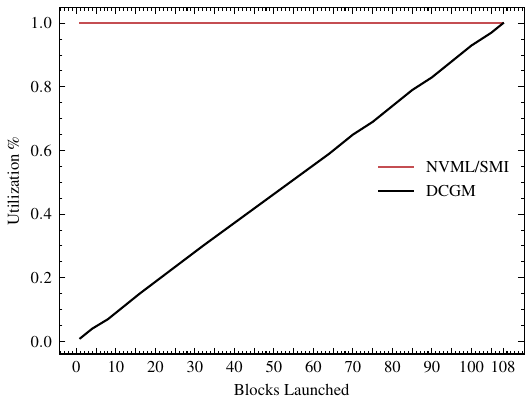}
                \caption{Running an infinitely looping kernel on a variable number of SMs. The popularly used NVML/SMI metric may significantly over-estimate the actual fraction of GPU SMs used.}
                \label{fig:sm_active_smi_comparison:example_1}
        \end{subfigure}
        \hfill
        \begin{subfigure}[t]{0.48\textwidth}
                \includegraphics[width=\textwidth]{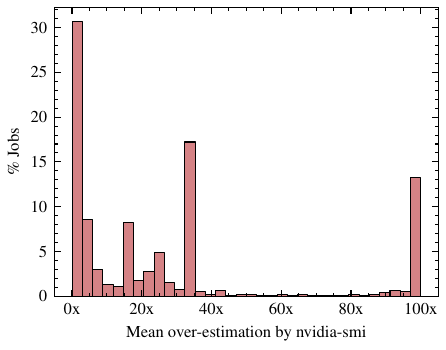}
                \caption{Distribution of per-job utilization overestimates reported by \texttt{nvidia-smi} compared to the DCGM metric.}
                \label{fig:sm_active_smi_comparison:distribution}
        \end{subfigure}
        \Description[Fig a. x-axis is num. blocks launched; y-axis is percent SM utilization; Fig. b. x-axis is mean-over-estimation by nvidia-smi across cluster; y-axis is percentage of jobs.]{}
        \caption{Comparisons of nvidia-smi and actual (reported by DCGM) GPU SM utilization.}
        \label{fig:sm_active_smi_comparison}
\end{figure}

Figure~\ref{fig:sm_active_smi_comparison:example_1} illustrates the discrepancy between \texttt{nvidia-smi} reported utilization and actual GPU usage by showing an example of an infinite-loop kernel (20 seconds) launched with a variable number of blocks (on an A100 with 108 SMs). DCGM-reported utilization gives an accurate view of intra-GPU spatial utilization. After deploying the monitoring system we compared the real-world differences. Figure~\ref{fig:sm_active_smi_comparison:distribution} highlights the ratio of average utilization reported by the two approaches. This systematic discrepancy has significant implications for cluster efficiency analysis and scheduling decisions, especially in scenarios involving multitasking.

\subsection{Design, Implementation and Deployment}

To meet the design requirements, we design a monitoring tool to run on each host node, recording its data to a database. The main loop of the tool queries DCGM along with \texttt{/proc} to retrieve metrics of each attached GPU at a given frequency, writes the returned results to a buffer, and then goes to sleep. When the buffer is full, the tool writes its content to its dedicated DB. At a much lower frequency (e.g., every hour), the tool queries the GPU cluster job scheduler (currently SLURM) for job information and writes to a separate table in the DB.

To achieve the low-overhead and efficiency goals, we developed the monitoring tool as a standalone executable, written in a few hundred lines of C. We set up a SQLite DB for each host node to avoid any synchronization among multiple hosts. This design decision makes the monitoring tool robust (immune to host failures), though it requires some work to aggregate the DB data offline for data analysis.

For our data collection, the sample period is configured to 100~ms, aligning with modern OS preemptive scheduling time slices. The local buffer size is configured to 600 samples to amortize the overhead of writing to DB. With such configurations, the buffer consumes $<$ 2 MB. On a host node with 4 GPUs, the sample loop takes about 800 microseconds. Every 60 seconds, the tool spends about 60~ms to write $\approx$ 24k rows totaling 1 MB to DB. Thus, the overhead is $\approx$ 0.9\% on a single CPU core. The total DB storage is about 1.5~GB/day per node. The overhead on GPUs is negligible. Because of its minimal overhead and robustness, the university IT administrators allowed us to deploy the monitoring system on a production research GPU cluster.

Our tool is open source and available on \href{https://github.com/als244/gpu_server_monitoring}{GitHub}.

\section{Job Characteristics}

\subsection*{Job Distribution by Department}

During the 10-day period, 7,399 jobs were submitted by 150 users from 12 departments. Table~\ref{table:job_by_department} summarizes the distribution of jobs by department (inferred from users' group affiliations) and the proportion of GPU time reserved by each department, providing insights into the cluster's diverse usage patterns.
Computationally intensive fields such as Chemistry and Chemical \& Biological Engineering dominate GPU usage, while departments like Neuroscience and Sociology, though contributing fewer jobs, add to the cluster's workload diversity.

\begin{table}[!htbp]
    \centering
    \begin{tabular}{||c | c | c ||}
        \hline
        Department       & \# of Jobs  & Prop. of Reserved GPU-Time  \\ [0.5ex]
        \hline\hline
        Astronomy        & 136 & 5.6\%       \\
        \hline
        Chem \& Bio Eng. & 2184 & 22.7\%       \\
        \hline
        Chemistry        & 2837 & 38.2\%      \\
        \hline
        Comp. Science    & 306  & 4.9\%      \\
        \hline
        Elect. Eng.      & 227  & 4.6\%      \\
        \hline
        Genomics         & 119 & 0.9\%        \\
        \hline
        Mech. Eng        & 241  & 11.6\%      \\
        \hline
        Neuroscience     & 385  & 3.4\%      \\
        \hline
        Op. Research     & 181  & 0.7\%      \\
        \hline
        Physics          & 569  & 3.1\%      \\
        \hline
        Psychology       & 28   & 1.0\%      \\
        \hline
        Sociology        & 90   & 1.6\%      \\
        \hline
    \end{tabular}
    \vspace{1em}
    \caption{Breakdown of jobs in the dataset by department. Departments which reserved $\leq 0.5\%$ of overall GPU-time are omitted.}
    \label{table:job_by_department}
\end{table}

\FloatBarrier
\subsection*{Job Duration and Categorization}

\begin{figure}[!htbp]
    \centering
    \includegraphics[width=0.45\linewidth]{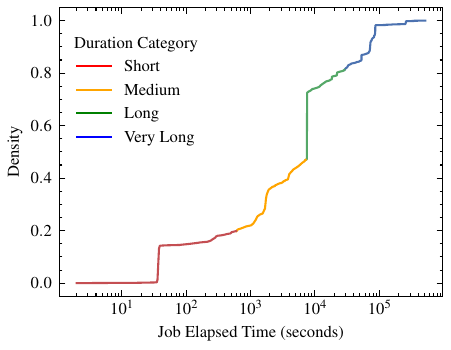}
    \caption{CDF of job elapsed time ($N=7,399$). Colored segments correspond to our different duration categories.}
    \label{fig:characterization_duration}
\end{figure}

Figure~\ref{fig:characterization_duration} presents the cumulative distribution function (CDF) of job elapsed time, which we categorized into four groups for easier analysis:
\begin{itemize}
    \item Short: $<$600 seconds (10 minutes)
    \item Medium: 600--7200 seconds (10 minutes--2 hours)
    \item Long: 7200--28800 seconds (2--8 hours)
    \item Very Long: $>$28800 seconds (8 hours)
\end{itemize}

Each category represents roughly 25\% of the jobs. The CDF reveals discrete jumps, suggesting that jobs are often submitted with specific common time limits, possibly reflecting users' reliance on predefined configurations.

\subsection*{GPU Resource Requests}

\begin{figure}[!htbp]
    \centering
    \begin{subfigure}[t]{0.48\textwidth}
            \centering

            \includegraphics[width=\textwidth]{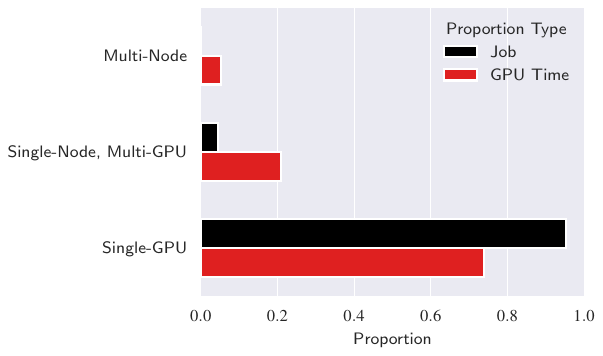}
            \caption{}
            \label{fig:job_time_props}
    \end{subfigure}
    \hfill
    \begin{subfigure}[t]{0.48\textwidth}
            \centering

            \includegraphics[width=\linewidth]{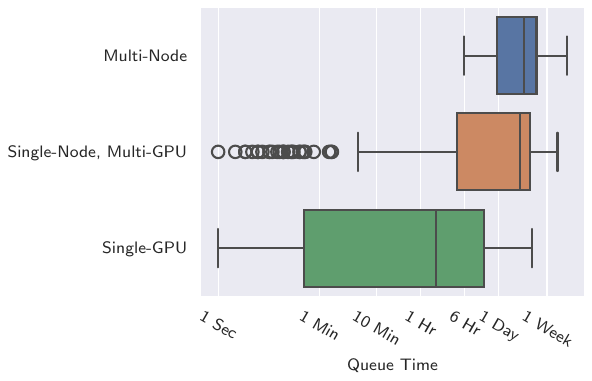}
            \caption{}
            \label{fig:queue_dist}
    \end{subfigure}
    \Description[]{Fig a. shows proportion of jobs and total GPU time, split by single-GPU, multi-GPU, and multi-node jobs. Most are single-GPU. Figure b. shows queue times for jobs, split by the same categories. Queue times are longest for multi-node jobs, followed by multi-GPU, followed by single-GPU. }
    \caption{Classifying by Requested Resources. $(a)$ Proportions of all jobs and total allocated GPU-Time and $(b)$ Corresponding time spent in queue. We find multi-node and multi-GPU jobs face longer queue times than single-GPU jobs. Moreover, peak queue times even for single-GPU jobs can exceed several days.}
    \label{fig:classification_breakdown}
\end{figure}

Figure~\ref{fig:job_time_props} illustrates the distribution of jobs based on the number of GPUs and nodes requested at submission. A significant majority (95\%) of jobs request a single GPU, with only a small fraction using multiple GPUs. However, multi-GPU and multi-node jobs take a disproportionately large amount of the GPU time on the cluster. These findings indicate that most workloads have limited parallelization in our dataset, potentially exacerbating the under-utilization of I/O and network bandwidth. This trend could be attributed to the complexity of parallelizing workloads across multiple GPUs or the longer queue times associated with multi-GPU jobs. This hypothesis is supported by Figure~\ref{fig:queue_dist}; jobs which request multi-GPU or multi-node configurations have significantly longer queue times on average, potentially encouraging frustrated users to prefer issuing single-GPU jobs instead.

The predominance of single-GPU jobs appears to be a factor contributing to the underutilization of the GPU cluster, including its network and I/O resources. The high timeout rate highlights a limitation in the current job-scheduling paradigm, requiring users to carefully estimate and request appropriate resources for their jobs.

\section{Our Findings}

\paragraph{\textbf{Significant Queuing.}} To better understand the cluster dynamics we categorize jobs into three groups: single-GPU, single-node with multiple GPUs, and multi-node. Figure \ref{fig:queue_dist} portrays the distributions of queue-times for each class; it is clear that requesting more resources results in longer queuing. The gap between requesting $1$ GPU vs. multiple is significant. The median queue time for Single-GPU jobs was already nearly $2$ hours. However, 2 hours does not seem long compared to the 2-\textit{day} median queue time for multi-GPU requests. Given these dynamics, it makes sense that the vast majority of jobs only request a single-GPU.

\begin{figure}[!htbp]
    \centering
    \begin{subfigure}[t]{0.48\textwidth}
        \centering
        \includegraphics[width=\textwidth]{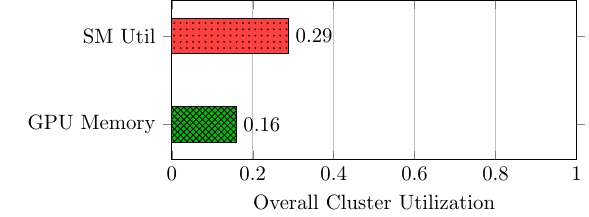}
        \caption{GPU Metrics}
        \label{fig:cluster_gpu_avgs}
    \end{subfigure}
    \hfill
    \begin{subfigure}[t]{0.48\textwidth}
        \centering
        \includegraphics[width=\textwidth]{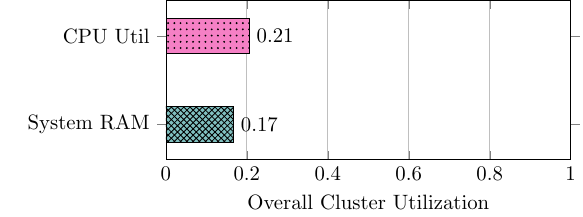}
        \caption{CPU Metrics}
        \label{fig:cluster_cpu_avgs}
        \end{subfigure}
    \Description[Cluster utilization metrics; Mean SM Util is 29 percent, and GPU memory usage is 16 percent across all jobs in cluster. Overall CPU utilization is 21 percent and host RAM usage is 17 percent]{}
    \caption{Aggregate cluster utilization. Averaged over the entire cluster throughout the 10-day sample. Overall, we find both GPU and CPU resources to be under-utilized across the cluster.}
    \label{fig:aggregate_cluster_utilization}
\end{figure}

\paragraph{\textbf{Under-utilization.}} If jobs are sitting in a queue for so long, we might expect high overall cluster utilization. Figure \ref{fig:aggregate_cluster_utilization} displays the aggregate cluster utilization over this 10-day window. Despite the inflow of jobs, cluster-utilization remains low with an overall GPU compute utilization of 29\% and GPU-memory utilization of 16\%, with CPU resources also sitting largely unused. These aggregate statistics echo prior monitoring studies (see \S \ref{sec:related_works}).

We investigate factors contributing to the widespread under-utilization and examine opportunities for future system designs.

\FloatBarrier

\subsection{Significant Mismatch between Requested vs. Actual Durations}
\label{interface}

\begin{figure}[!htbp]
    \centering
     \includegraphics[width=.95\textwidth]{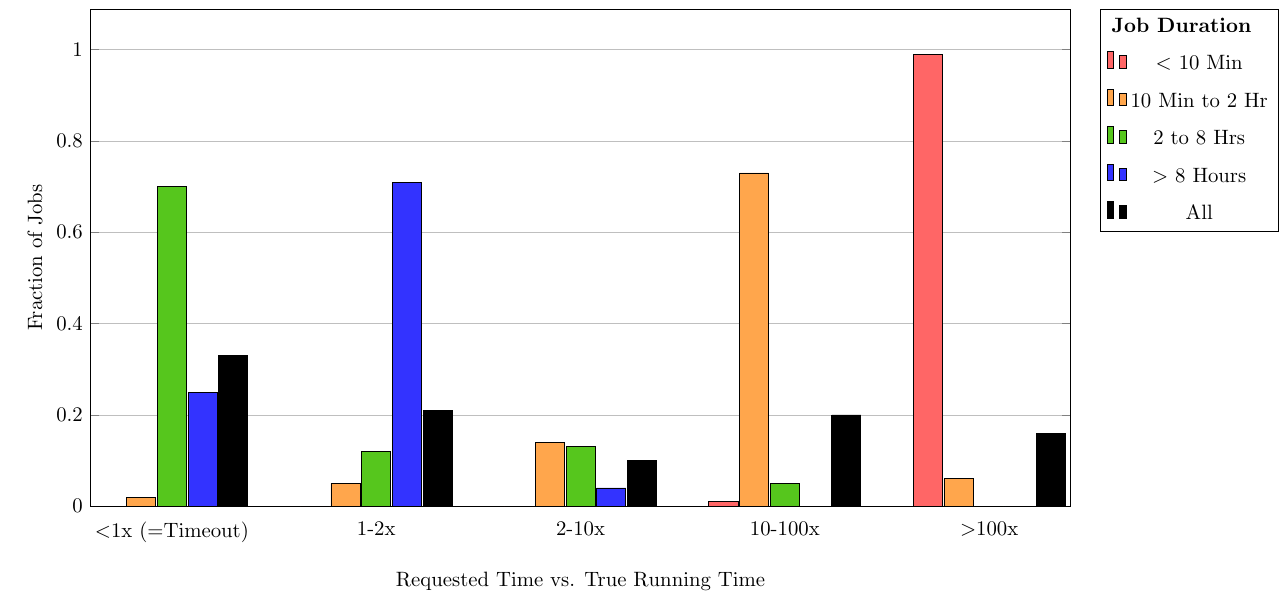}
    \caption{A significant portion of jobs over-request time. This is particularly troublesome for the longer jobs which present difficulties for scheduling algorithms as they are receiving inaccurate inputs. This can result in poor global decisions and thus excessive queuing.}
    \label{fig:over_req_ratio}
\end{figure}

We find there is a widespread mismatch between requested and actual time durations of the monitored jobs.
Figure~\ref{fig:over_req_ratio} illustrates the ratio of requested time to actual elapsed time for jobs, categorized by job duration, with about 80\% of jobs either underestimated or overestimated by a large factor. Only about 20\% of requested durations fall within a factor of two of the actual runtime, and a substantial percentage are off by more than an order of magnitude.

Table~\ref{tab:completion_states} shows the distribution of job completion states. While 60.7\% of jobs complete successfully, a notable {\bf 34.2\% time out}, and the remaining 5.1\% fail due to user cancellations, errors, or host memory constraints.
The high timeout rate suggests that many users underestimate their job runtimes, wasting the GPU time of prematurely terminated jobs, though some timeouts are intentional (e.g., interactive sessions or checkpointed jobs that run until their limit).

\begin{table}[!htbp]
    \centering
    \begin{tabular}{||c|c||}
        \hline
        Completion State & Percentage of Jobs \\
        \hline\hline
         Completed  &   60.70   \\ \hline
         Timeout    &   34.17   \\ \hline
         Failed     &   3.15    \\ \hline
         Canceled   &   1.99    \\ \hline
    \end{tabular}
    \caption{Completion states of jobs ($N=7,399$) on the cluster. A significant portion (34.17\%) of jobs reach their requested time limit. A smaller fraction are canceled by the user, or fail due to errors or host memory constraints.}
    \label{tab:completion_states}
\end{table}

Over-requesting resources also has adverse effects. From the users' perspective, overestimating job durations increases queue delays, as the scheduler must allocate larger time slots for these jobs.

An important set of jobs are essentially idle during their actual execution time periods. Table~\ref{tab:idle_jobs} depicts the breakdown of jobs with fully-idle GPU-usage by duration. The vast majority, 1,043 jobs, lasted less than a minute; of these 1,041 report successful completion, likely implying reservations for very small jobs or configuration errors where a GPU was accidentally requested.

\begin{table}[!htbp]
\centering
    \begin{tabular}{||c | c | c ||}
        \hline
        Duration         & Prop. of All Jobs & Prop. of Reserved GPU-Time \\
        \hline\hline
        $<$ 1 Minute       & 14.1\%   & 0.02\%   \\
        \hline
        [1 Minute, 10 Minutes) & 0.8\% & 0.001\%   \\
        \hline
        [10 Minutes, 2 Hours] & 2.2\%   & 0.44\% \\
        \hline
        [2 Hours, 8 Hours]    & 0.4\%    & 0.12\%  \\
        \hline
        $>$ 8 Hours     & 0.6\% & 2.15\%     \\
        \hline
    \end{tabular}
\vspace{1em}
\caption{Fully-Idle GPU Jobs}
\label{tab:idle_jobs}
\end{table}

A non-negligible number of jobs lasting multiple hours with fully-idle GPUs present a ripe opportunity for system designers. Out of all of the reserved GPU-time, 2.82\% was spent in fully idle jobs, with 2.15\% allocated as part of long-running ($>$ 8 hours), completely idle jobs. Whether it be user-error in accidental GPU requests, or optimistic reservations that ended up not being needed, it is clear that an intelligent cluster would detect and take advantage of idle resources, rather than locking them away.

\subsection*{Implications for Scheduling}

The widespread and drastic mismatch between requested and actual durations leads to poor scheduling decisions, causing low job completion rates, reduced cluster utilization, and increased queuing delays.

There are several significant implications:
\begin{itemize}
    \item \textit{Motivation for Avoiding User-Specified Time}:
  Our data reveal a severe and widespread mismatch between user-specified and actual durations, with one-third of jobs hitting their requested time limit. This discrepancy results in substantial waste of GPU resources and valuable human effort.

    \item \textit{Motivation for Automatic Runtime Predictions}: Since the effectiveness of schedulers depends on the accuracy of their input data, using historical data or learning-based approaches could help schedulers make better decisions.

    \item \textit{Motivation for \underline{Multitasking}}:
  Multitasking approaches can significantly enhance GPU utilization by running multiple tasks concurrently, effectively leveraging underutilized resources.
\end{itemize}

These implications motivate systems designers to explore new methods to manage GPU clusters.

\FloatBarrier

\subsection{Great Opportunities for Multitasking}
\label{multiplexing}

Most cluster management systems require users to reserve entire GPUs for fixed durations, effectively ``locking away'' resources that often remain underutilized. While some jobs fully saturate their GPUs, our data show that this is typically only for brief periods, if at all. Many reservations result in significant underutilization, raising the question of whether allocating entire GPUs on an all-or-nothing basis is the most effective approach, especially given the rising cost and demand for modern GPUs.

\subsection*{Low SM and Memory Utilizations}
\begin{figure}[!htbp]
    \centering
    \begin{subfigure}[t]{0.48\textwidth}
        \centering
        \includegraphics[width=\textwidth]{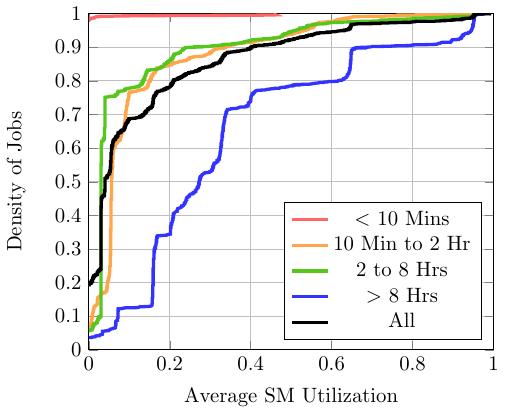}
        \caption{}
        \label{fig:avg_gpu_util}
    \end{subfigure}
    \hfill
    \begin{subfigure}[t]{0.48\textwidth}
        \centering
        \includegraphics[width=\textwidth]{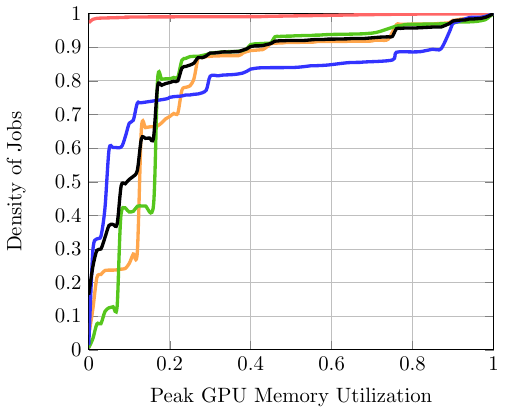}
        \caption{}
        \label{fig:peak_mem_usage}
    \end{subfigure}

    \caption{CDFs of Single-GPU Job Utilization Metrics, stratified by duration. (a) Average SM Utilization (b) Peak Memory Usage}
    \label{fig:resource_frag}
\end{figure}

To study GPU utilization,
we analyze the average Streaming Multiprocessor (SM) utilization and peak memory usage for single-GPU jobs, categorized by job duration. The results, illustrated in Figure~\ref{fig:resource_frag}, show significant underutilization:

\begin{itemize}
    \item 50\% of jobs sustain an average SM utilization of under 5\%, and
    \item 75\% of jobs have a peak memory usage of less than 20\% and an average SM usage of under 20\%.
\end{itemize}

These observations suggest that most jobs are far from fully utilizing GPU capabilities, a trend likely to worsen as per-GPU performance grows faster than typical workload demands.

Efficient memory sharing is often a critical constraint for multitasking, as one task running out of memory can disrupt other tasks. Figure~\ref{fig:peak_mem_usage} shows that only 5\% of jobs ever reach 75\% of GPU memory capacity, and among long-running jobs (over 8 hours), just 10\% exceed 90\%. These results suggest that while some jobs may have large working sets, most leave ample memory headroom, making multitasking feasible.

\begin{figure}[!htbp]
    \centering
    \begin{subfigure}[t]{0.48\textwidth}
        \centering
        \includegraphics[width=\textwidth]{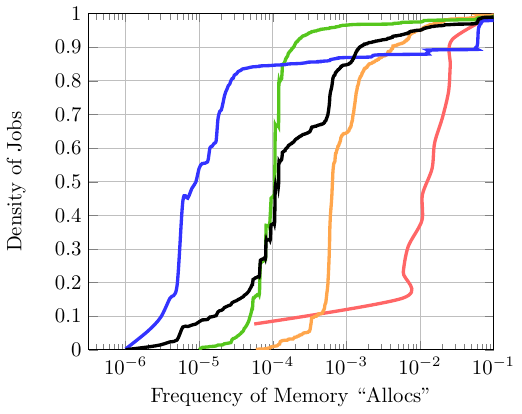}
        \caption{}
        \label{fig:alloc_frequency}
    \end{subfigure}
    \begin{subfigure}[t]{0.48\textwidth}
        \centering
        \includegraphics[width=\textwidth]{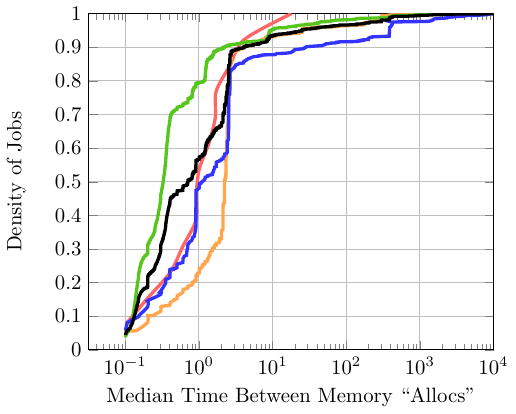}
        \caption{}
        \label{fig:time_between_allocs}
    \end{subfigure}
    \hfill
    \caption{Memory Stability: (a) Fraction of recorded samples where memory usage increased across all jobs. (b) Median time between samples where memory usage increased, across all jobs. Line colors correspond to the duration categories in Figure~\ref{fig:resource_frag}.}
    \label{fig:mem_stability}
\end{figure}

Figure~\ref{fig:mem_stability} provides additional insights into memory stability. Memory-volatility data (Figure~\ref{fig:alloc_frequency}) reveals how frequently memory usage increases during a job, while Figure~\ref{fig:time_between_allocs} shows the typical time gap between memory growth events. Many jobs exhibit relatively stable memory usage for extended periods, reducing the risk of memory churn in multitasking scenarios.

Since single-GPU jobs are not gang-scheduled, they are well-suited for multitasking. We have performed a separate analysis of their SM and memory utilization, as illustrated in Figure~\ref{fig:avg_gpu_util}. The figure highlights the following:

\begin{itemize}
    \item Only 10\% of single-GPU jobs achieve an average SM utilization exceeding 50\%.
    \item Among long-running jobs (over 8 hours), 33\% surpass this threshold, indicating slightly higher but still limited utilization.
    \item Over 70\% of single-GPU jobs utilize less than 20\% of memory capacity.
\end{itemize}
These findings align closely with the overall utilization trends, as most jobs in the workload are single-GPU jobs.

\FloatBarrier

\subsection*{Fine Temporal Granularity}

\begin{figure}[!htbp]
    \centering
    \begin{subfigure}[t]{.48\textwidth}
        \includegraphics[width=\linewidth,height=0.38\textheight,keepaspectratio]{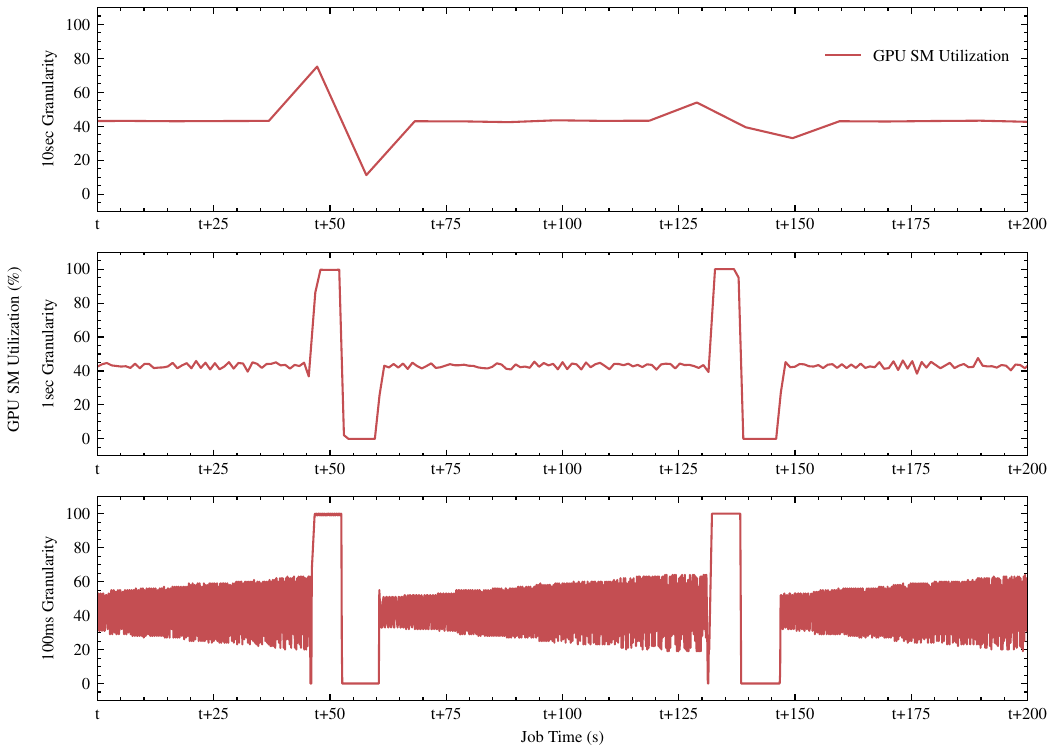}
        \caption{Comparison of querying GPU SM utilization at different sample rates. We see significant differences in observed program behavior as a result of the sampling rate; details such as the high frequency fluctuations and deep troughs in SM utilization are lost when sampling at lower rates.}
        \label{fig:sm_active_temporal_comparison}
    \end{subfigure}
    \hfill
    \begin{subfigure}[t]{.48\textwidth}
        \includegraphics[width=\linewidth,height=0.38\textheight,keepaspectratio]{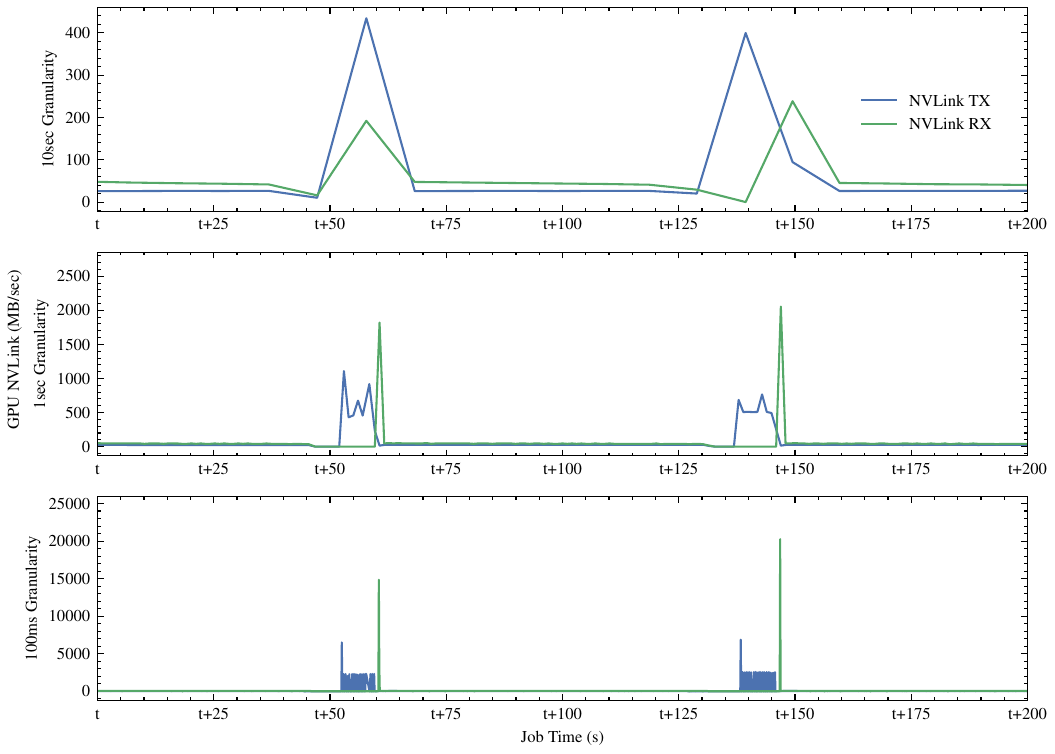}
        \caption{Comparison of querying NVLink throughput at different sample rates. Note the differing y-axis. As granularity increases, we gain more insight into the burstiness of NVLink traffic.}
        \label{fig:nvlink_temporal_comparison}
    \end{subfigure}
    \caption{Comparisons of querying GPU SM utilization and NVLink throughput at different sample rates.}
    \label{fig:temporal_comparison}
\end{figure}

\begin{figure}[!htbp]
    \centering
    \includegraphics[width=0.45\linewidth]{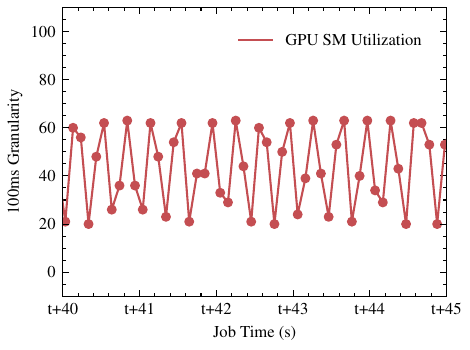}
    \caption{Zoomed-in view of Figure~\ref{fig:sm_active_temporal_comparison}}
    \label{fig:zoomed_100msec_trace_sm}
\end{figure}

The temporal granularity of GPU utilization sampling plays a crucial role in identifying underutilized intervals. Figure~\ref{fig:temporal_comparison} compares sampling rates of 10 seconds, 1 second, and 100 milliseconds for both SM usage (Figure~\ref{fig:sm_active_temporal_comparison}) and NVLink throughput (Figure~\ref{fig:nvlink_temporal_comparison}). Coarser sampling masks significant fluctuations and bursty behavior, potentially misleading schedulers into overestimating GPU load. In contrast, 100~ms sampling, as shown in the zoomed-in trace in Figure~\ref{fig:zoomed_100msec_trace_sm}, captures the more accurate peaks and troughs in SM utilization. Similarly, NVLink traffic reveals highly bursty patterns at fine-grained timescales.

These data demonstrate the importance of granular telemetry when considering optimal resource allocation. A scheduler can take advantage of fine-grained utilization data to dynamically enqueue additional tasks onto a GPU without waiting for the current job to complete.

\noindent\begin{minipage}{\linewidth}
\subsection*{Implications for Multitasking}
Our data expose the severe underutilization of GPU SMs and memory, revealing great opportunities for multitasking on GPUs. There are several implications:
\begin{itemize}
    \item \textit{Motivation for multitasking}: Dynamic multitasking could unlock significant available resources, particularly since over 70\% of jobs consume less than 20\% of SMs and memory.

    \item \textit{Opportunity to Pack Single-GPU Jobs}: Single-GPU jobs are good candidates for multiplexing as they are not gang-scheduled and do not incur inter-GPU traffic. These jobs account for a substantial fraction of aggregate cluster usage.

    \item \textit{Motivation for Using Fine-Grained Time Slices}: Fine-grained preemptive scheduling time slices such as 100--200\,ms, on the order of the round-robin time slices OS schedulers use for CPUs, give more opportunities to improve GPU utilization.
\end{itemize}
\end{minipage}

Nvidia's Multi-Instance GPU (MIG)~\cite{nvidia_mig}, Multi-Process Service (MPS)~\cite{nvidia_mps}, and Unified Virtual Memory (UVM)~\cite{nvidia_uvm} are advancing toward greater maturity. These mechanisms offer promising opportunities to explore multitasking and reclaim underutilized GPU resources.

\FloatBarrier

\subsection{Highly Periodic, and Predictable, Communication Patterns}

While most of the jobs on this cluster are isolated single-GPU jobs, Figure~\ref{fig:job_time_props} indicates that 26\% of allocated GPU time was for parallel workloads, and we presume this quantity to be significantly higher for AI and HPC dedicated clusters. The nature of parallel workloads and their accompanying synchronization poses a significant challenge to multitasking. Unintended interference can have compounding straggler effects~\cite{wu2024falconpinpointingmitigatingstragglers}.

To determine the possibility of interleaving tasks alongside parallel workloads we investigated communication patterns of Multi-GPU workloads at a link utilization granularity. Given the large number of jobs, we present three representative examples.

\subsection*{Single-Node Patterns}

\subsubsection*{Neuroscience Job}

\begin{figure}[!htbp]
     \begin{subfigure}[t]{0.5\textwidth}
        \centering
        \includegraphics[width=\textwidth]{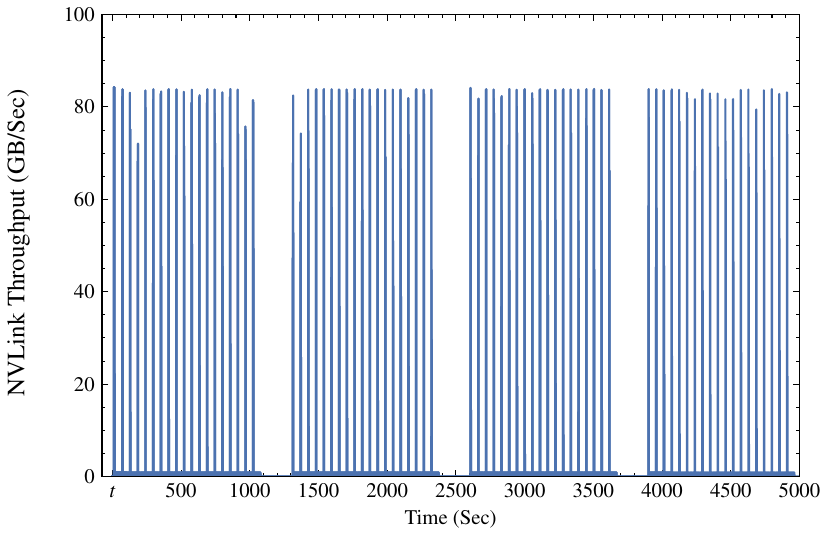}
        \caption{$\approx$ 17-Minute Phase Periodicity}
        \label{fig:neuro_wide}
    \end{subfigure}
    \hfill
     \begin{subfigure}[t]{0.43\textwidth}
        \centering
        \includegraphics[width=\textwidth]{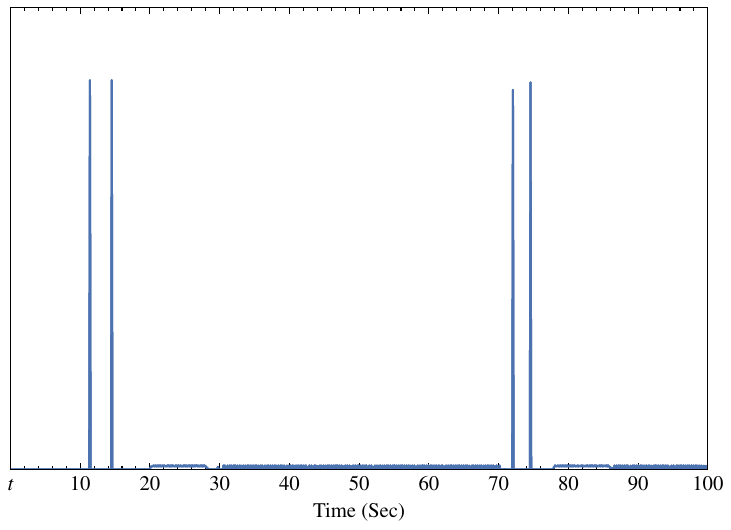}
        \caption{Localized, Per-Minute, Bursts}
        \label{fig:neuro_zoomed}
    \end{subfigure}

    \caption{Single-Node Neuroscience Job}
    \label{fig:neuro_trace}
\end{figure}

Figures~\ref{fig:neuro_wide} and \ref{fig:neuro_zoomed} illustrate NVLink usage in a parallel neuroscience workload using 4 GPUs within a single node. This job demonstrates a periodic communication phase of approximately 17 minutes, with additional bursts occurring roughly once per minute. The left panel (\figurename~\ref{fig:neuro_wide}) shows the consistent repetition of large-scale patterns throughout the job's runtime, while the right panel (\figurename~\ref{fig:neuro_zoomed}) zooms in on the finer-grained bursts. This structured periodicity indicates that higher-level software systems can estimate NVLink traffic given a small initial sample of metrics.

\subsubsection*{Psychology Job}
\begin{figure}[!htbp]
     \begin{subfigure}[t]{0.5\textwidth}
        \centering
        \includegraphics[width=\textwidth]{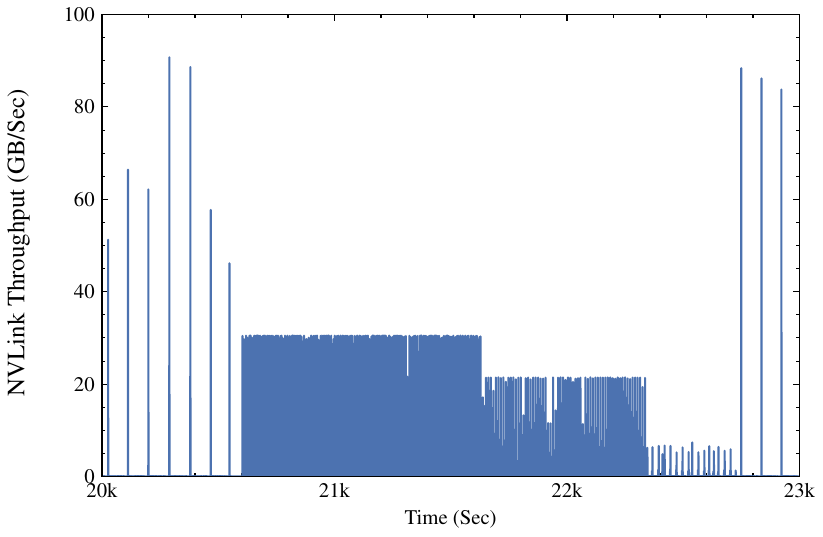}
        \caption{``Bursty'' Traffic at Two Different Rates}
        \label{fig:psyc_wide}
    \end{subfigure}
    \hfill
    \begin{subfigure}[t]{0.43\textwidth}
        \centering
        \includegraphics[width=\textwidth]{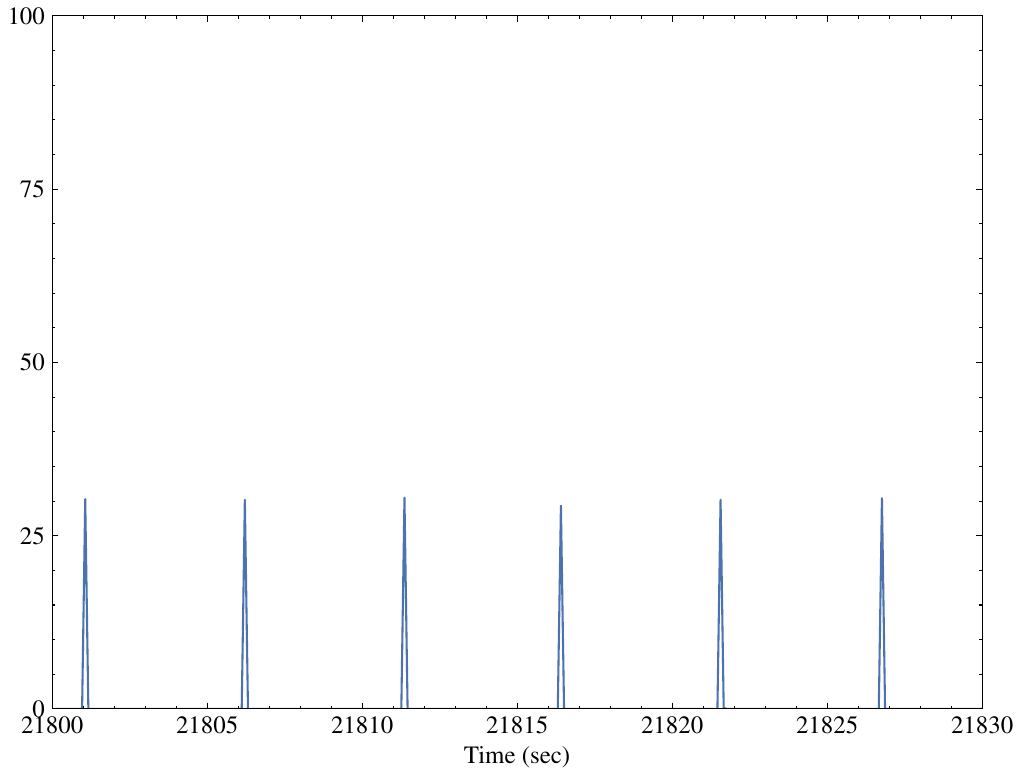}
        \caption{Focusing in on denser rectangular regions in the center of \ref{fig:psyc_wide} still reveals sparsity and periodicity --- once per 10 seconds vs. once per 60 seconds (surrounding spikes in \ref{fig:psyc_wide})}
        \label{fig:psyc_zoomed}
    \end{subfigure}
    \caption{Single-Node Psychology Job}
    \label{fig:pysc_trace}
\end{figure}

Figures~\ref{fig:psyc_wide} and \ref{fig:psyc_zoomed} depict a psychology workload, also using 4 GPUs within a single node. This job exhibits mixed burst patterns, with bursts occurring at intervals of approximately 10 seconds and 60 seconds, respectively. While these intervals differ from the neuroscience example, the repetitive nature of NVLink traffic remains consistent, reflecting the structured communication demands of the application.

\FloatBarrier

\subsection*{Comprehensive Picture}

\begin{figure}[!htbp]
        \centering
        \includegraphics[width=\linewidth,height=0.78\textheight,keepaspectratio]{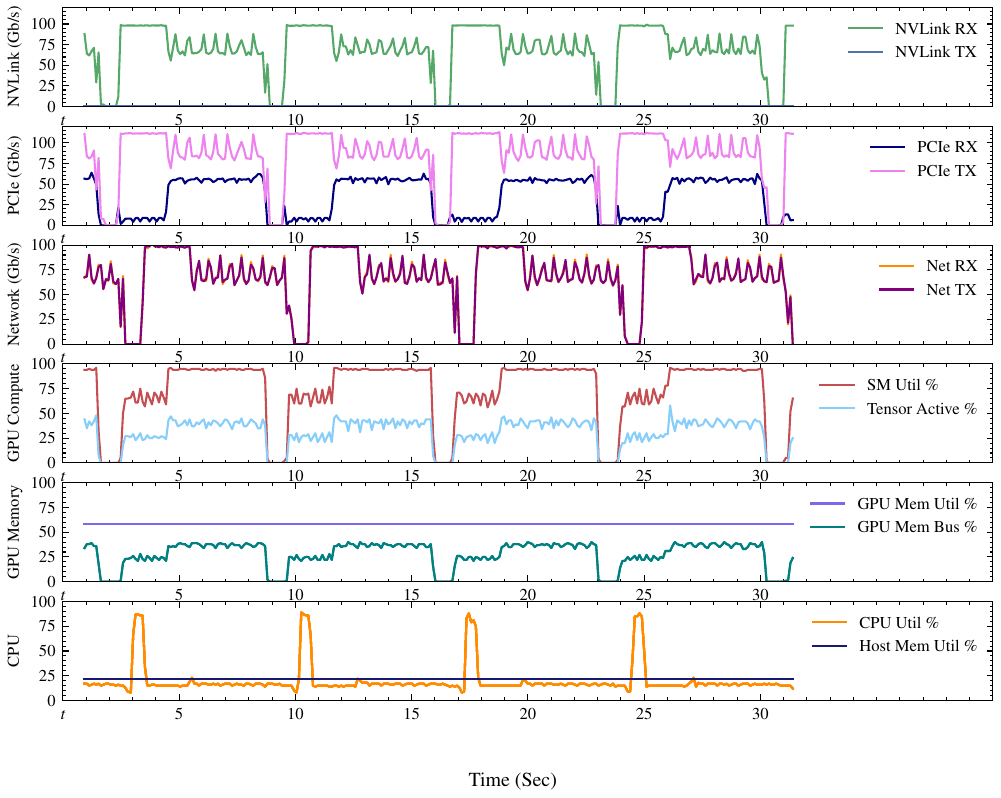}
    \caption{Slice from 4-Node, 16-GPU workload: Periodicity across all resources \& frequent pockets of SM idleness (4th row, red)}
    \label{fig:comp_view}
\end{figure}

Periodicity among all resource dimensions can be seen in Figure~\ref{fig:comp_view}, captured from a 4-Node, 16-GPU EE job, likely doing AI training; this pattern repeats for 12 hours with almost no noise. We can infer communication behavior through a combination of traffic metrics. The network metrics come from the node's singular InfiniBand NIC, while the interconnect metrics are for a specific GPU within the node. The PCIe TX/RX metric traffic includes bytes sent to/received from the network card during RDMA in addition to bytes sent/received from system memory. Note that interconnect graphs are in network units (Gb/s).

We see that this job is clearly bottlenecked by network bandwidth (purple/orange --- the TX/RX lines are nearly identical), which reaches full saturation of $\approx$ 100~Gb/s for 2-second windows. In this scenario, both NVLink and PCIe are constrained to the network limit.

We can deduce the selected GPU in Figure~\ref{fig:comp_view} is continually transmitting over the network as the PCIe TX readings (in pink, second plot) align with the node's network TX readings. It is not transmitting to other intra-node GPUs (0 NVLink TX traffic from this selected GPU). The PCIe RX throughput implies inbound network traffic is split among multiple GPUs in this system. Other GPUs initially receive traffic (flat blue PCIe RX line) and then this GPU receives a partial share. These readings agree with the intra-node NVLink RX traffic this GPU receives (top plot in green) as it first is reading at the network limit and then drops off when it is receiving some data directly from the NIC via RDMA.

We highlight that GPU SM Utilization (in red) fluctuates in lock-step with the overall communication pattern. Fully-idle GPUs result from what we presume to be global synchronization. In this example for every 7-second period, 1 second is spent idle. Accounting for 16 GPUs and a 12-hour duration, this implies around 21 hours of available GPU-computation that can be packed into these troughs.

The motif of small pockets of SM idleness, occurring at frequent and regular intervals, is the common case for many network-bound distributed GPU applications. Despite its highly parallel nature, this type of job is a valid candidate for multitasking. GPU memory usage is shown staying flat at 50\% (it does not change for hours) in lavender (second from bottom), leaving ample space for other jobs' data.

\subsection*{Implications for Distributed Jobs}

Our analysis underscores the importance of high GPU-interconnect and network bandwidth. Single-Node jobs can exhibit full NVLink saturation during blocking communication phases. Figure~\ref{fig:temporal_comparison} highlights how insufficient interconnect performance can degrade throughput during communication-intensive phases. Distributed jobs can be bottlenecked by network bandwidth and fail to saturate high-BW NVLinks and ultimately cause SM idle periods. The analysis indicates:

\begin{itemize}
    \item \textit{Opportunity for Accurate Traffic Predictions}: Repetitive communication patterns present an opportunity for accurate traffic prediction. Scheduling tasks in an ``out-of-phase'' manner could help minimize contention and maintain consistent throughput.
    \item \textit{Opportunity to Address Synchronization-Induced Idle Periods}:
   Transparent multitasking alongside long-running distributed jobs can significantly improve cluster utilization, especially in bandwidth-limited environments where communication delays often cause extended GPU idleness.
\end{itemize}
The general challenge to take advantage of these opportunities is to minimize interference during the bursty communication periods.

\section{Related Work}
\label{sec:related_works}

\begin{table}[!htbp]
    \centering
    \small
    \setlength{\tabcolsep}{4pt}
    \begin{tabularx}{\textwidth}{||>{\raggedright\arraybackslash}p{0.28\textwidth}|>{\centering\arraybackslash}p{0.16\textwidth}|>{\centering\arraybackslash}p{0.18\textwidth}|>{\raggedright\arraybackslash}X||}
        \hline
        Prior Work &
        Spatial \mbox{Granularity} &
        Temporal Granularity &
        Studied Workloads \\
        \hline\hline
        Philly \cite{jeon_analysis_2019} & Coarse & 100 seconds & Deep Learning\\
        \hline
        AntMan \cite{xiao_antman_2020} & Coarse & \(\approx\) 1 second\textsuperscript{1} & Deep Learning\\
        \hline
        Helios \cite{hu_characterization_2021} & Coarse & 100 seconds & Deep Learning \\
        \hline
        MIT SuperCloud \cite{samsi_supercloud_2021} & Coarse\textsuperscript{4} & 100 ms & Mixed (incl.\ DL)\\
        \hline
        \citet{weng_mlaas_2022} & Coarse & 15 seconds & Deep Learning\\
        \hline
        Moneo \cite{jiang_moneo_2022} & Fine & Variable\textsuperscript{2} & Individual DL Training Jobs\textsuperscript{3}\\
        \hline
        \textbf{Ours} & \textbf{Fine} & \textbf{100 ms} & \textbf{Mixed (12 disciplines)} \\
        \hline
    \end{tabularx}
    \vspace{0.5em}
    \begin{minipage}{0.95\textwidth}
    \footnotesize
    \textsuperscript{1}Inferred from time-series traces in \citet{xiao_antman_2020} and confirmed with authors.\\
    \textsuperscript{2}Supports DCGM monitoring, but dynamically reduces sampling rate.\\
    \textsuperscript{3}Does not evaluate cluster-wide utilization; studies 3 specific workloads.\\
    \textsuperscript{4}Samples \texttt{nvidia-smi} utilization, which does not capture fractional SM usage (\S\ref{sec:monitoring}).
    \end{minipage}
    \vspace{0.5em}
    \caption{Comparison to several prior works studying cluster utilization.}
    \label{tab:related_work_comparison}
\end{table}

\paragraph{GPU Monitoring APIs}
Despite growing awareness of underutilization, few production systems have deployed sub-second sampling to characterize bursty or partial GPU usage. We summarize key differences between our work and prior monitoring work in Table~\ref{tab:related_work_comparison}. The DCGM API supports high-resolution telemetry (e.g., SM Active, SM Occupancy, NVLink, PCIe) at 100--200\,ms intervals, but it is rarely used at scale. The MIT SuperCloud dataset~\cite{samsi_supercloud_2021} also records GPU time series every 100~ms, but through \texttt{nvidia-smi}, whose utilization metric can overstate SM usage by orders of magnitude (\S\ref{sec:monitoring}). To our knowledge, our work is among the first to collect DCGM's SM-level metrics at 100~ms across an entire multi-disciplinary research cluster. This approach reveals patterns---such as frequent GPU idleness, partial SM occupancy, and bursty interconnect usage---that remain hidden when sampling at intervals of one second or longer.

\paragraph{GPU Cluster Monitoring} Monitoring GPU-based clusters has been a significant topic of research, with numerous tools and frameworks designed to track resource usage. One of the earliest and most broadly adopted systems is \textit{Ganglia}~\cite{massie_ganglia_2004}, which provides cluster-wide resource utilization metrics via a hierarchical design. Although Ganglia can be extended to expose GPU statistics (e.g., through an NVIDIA GPU monitoring module), it typically leverages the NVIDIA Management Library (NVML)~\cite{nvidia_nvml}, offering only coarse-grained GPU usage data. Consequently, Ganglia-based monitoring lacks sub-second resolution and may miss rapid fluctuations in GPU compute, memory, or interconnect utilization. We instead designed our own monitoring solution to have more precise control over monitoring and sampling parameters.

We are not the first to champion fine-grained, cluster-wide GPU monitoring; Moneo~\cite{jiang_moneo_2022} is a tool developed by Microsoft which also adopts fine-grained monitoring of GPU metrics. While Jiang et al.\ propose capturing fine-grained metrics, they only study a limited selection of hand-picked deep learning training jobs. However, despite the availability of such tools, they remain unpopular for use on HPC research clusters; a recent study by \citet{weakley_monitoring_2025} surveys GPU monitoring practices in HPC clusters, finding that the newer NVIDIA Data Center GPU Manager (DCGM) API~\cite{nvidia_dcgm}---which supports sub-second sampling---remains under-adopted. They attribute this to limited compatibility with older GPUs and the overhead of integrating DCGM with existing monitoring systems. We hope that our work serves to raise broader awareness.

In practice, the predominant monitoring infrastructure consists of deploying a container-based solution composed of Nvidia's \verb+dcgm_exporter+ (for GPU metrics) and the Prometheus monitoring framework~\cite{prometheus} (for CPU-side metrics and a time-series DB). This setup is targeted for Kubernetes-based clusters and is non-trivial to set up and control. Our monitoring tool can be deployed on any bare-metal Nvidia Data-Center GPU server (without dependencies) and is trivial to start/stop from the command line (start/kill the daemon process).

\paragraph{Prior Characterization Studies} Several large-scale studies have examined GPU cluster utilization to identify inefficiencies and guide improvements in resource management; we provide a high-level summary of the differences with our work in Table~\ref{tab:related_work_comparison}. \textit{Philly}~\cite{jeon_analysis_2019} from Microsoft was among the first to report considerable underutilization in a large-scale multi-tenant GPU cluster dedicated mostly to deep learning. Philly relied on NVML-based sampling at 100-second intervals, limiting its ability to capture sub-second bursts or fractional GPU usage. While underutilization in Philly was largely attributed to network constraints and gang scheduling, our results demonstrate that, even on nodes running a single job, significant idle periods occur that only become apparent when sampling at finer granularity.
\textit{Helios}~\cite{hu_characterization_2021} took a closer look at multiple production clusters running various GPU workloads. Like Philly, Helios employed NVML for coarse-grained (100-second) sampling and did not measure fractional SM utilization. Although Helios confirmed the prevalence of low GPU utilization and long queue times, it left open questions about the root causes of underutilization within each job's runtime. More recently, \citet{weng_mlaas_2022} characterize utilization in a production cluster which supports time-multiplexed sharing of GPUs between jobs, demonstrating low contention for SM resources in their analysis. While a strong preliminary step towards improving utilization, our results show that significant fractions of GPU SM resources remain unutilized by jobs; as time-slicing alone cannot allow for the sharing of these resources, we feel fine-grained monitoring of utilization reveals greater opportunities for multiplexing. Our sub-second analysis highlights internal idle phases and scheduling opportunities that standard job-level metrics cannot readily detect.

\paragraph{GPU Multitasking and Scheduling Approaches} Many researchers have explored the idea of workload-independent GPU sharing~\cite{adriaens_case_2012,tanasic_enabling_2014,yun_liang_efficient_2015,park_chimera_2015,chen_effisha_2017,wang_simultaneous_2016, mahajan_themis_2020, sedighi_efficient_2024}. Much of recent work regarding GPU-multitasking has concentrated on machine-learning focused workloads. \textit{Gandiva}~\cite{xiao_gandiva_2018} considers time-slicing shared GPU resources between multiple jobs. \textit{AntMan}~\cite{xiao_antman_2020} introduces a software layer to multiplex GPU compute and memory resources, improving overall cluster usage. Like prior work, AntMan's monitoring relies on NVML at roughly 1\,s intervals, thus it still may miss quickly shifting utilization patterns. Nevertheless, its reported improvements underline the potential of GPU multi-tenancy---a conclusion our results reinforce. By measuring SM usage at sub-second granularity, we demonstrate that jobs often leave enough unused SM capacity for other processes to co-locate effectively.

More recently, \textit{Orion}~\cite{strati_orion_2024} proposes a method for fine-grained sharing of GPU workloads on a single node, showing that performant and interference-aware multiplexing of GPU resources is possible. However, Orion relies on GPU utilization profiles collected offline, limiting its applicability to general user applications. We believe there is room for systems based on dynamic, real-time profiling of workloads to allow for flexible, application-agnostic sharing of GPUs.
Based on the Helios dataset, \citet{hu_characterization_2021} proposed a ``Quasi-Shortest-Service-First (QSSF)'' scheduler, which uses historical job records to predict runtime and reduce queuing for large multi-GPU jobs. However, QSSF assumes relative stability in job behavior and easy access to training data, assumptions that may not hold in research environments where development and debugging are common.
Our findings suggest that more adaptive scheduling strategies, fed by fine-grained performance metrics, could mitigate these shortcomings.

\section{Limitations}

\paragraph{Sampling Rate} We used the highest sampling frequency, 10 Hz, offered by DCGM \cite{nvidia_dcgm}; however, we would have preferred to collect data at even higher rates if the API had allowed it, given our low-overhead monitoring tool supports rates 10--100$\times$ faster. While our temporal resolution exceeds that of prior monitoring studies, interconnect and network measurements are still aggregated over 100\,ms windows, which may smooth out short-lived communication spikes and obscure burstiness. Moreover, individual GPU kernels often execute on a timescale of microseconds to single-digit milliseconds. Profiling tools such as Nvidia Nsight Systems (\verb|nsys|) can capture this data (as a background process), but are not applicable for cluster-scale monitoring due to job degradation.

\paragraph{Duration of Study} Although we ran the monitoring tool for several months, we only retrieved precise mappings of Job ID to GPU ID(s) for the final 10 days. Rather than making potentially inaccurate inferences about which jobs ran on which devices, we limited our analysis to the clean, fully instrumented dataset.

\paragraph{Job Types and Diversity} Because most jobs in the cluster requested only a single GPU, we collected fewer samples of multi-GPU workloads (343 in total) and of these only 7 multi-node workloads. Our insight into the frequencies and behaviors of domain-specific workloads (e.g. ML or HPC workloads) is limited; our analysis aggregates data from all departments.

\paragraph{Cluster Hardware and Heterogeneity} The cluster included two GPU types (A100 PCIe 40~GB and A100 SXM4 80~GB). We combined them into a single dataset for simplicity, as we report utilization in percentages. However, these GPUs differ in their interconnect bandwidth (PCIe vs.\ NVLink). Because most jobs were single-node, the impact of these differences on our overall findings was small. Nonetheless, future work should investigate how heterogeneous hardware and network capabilities affect cluster utilization and communication patterns.

\section{Conclusion}
This paper presents a fine-grained analysis of a university research GPU cluster through a lightweight monitoring tool operating at 100~ms granularity. Our study reveals three key findings: (1) severe underutilization, with over 70\% of jobs using less than 20\% of GPU memory and 50\% of jobs averaging below 5\% SM utilization; (2) pervasive mismatch between requested and actual job durations, affecting 80\% of jobs and fundamentally limiting scheduler effectiveness; and (3) highly periodic communication patterns in multi-GPU jobs, suggesting opportunities for predictive resource allocation and compatible multiplexing.

These results, coupled with emerging technologies like MIG, MPS, and UVM, indicate significant potential for improving cluster efficiency through dynamic resource sharing and intelligent scheduling. By leveraging fine-grained monitoring data, system designers can develop multitasking solutions that maximize GPU utilization while maintaining performance isolation, potentially transforming how research GPU clusters are managed and utilized.

\begin{acks}
We thank Josko Plazonic, Jon Halverson, Curtis Hillegas, and the rest of the Princeton Research Computing team for helping deploy the monitoring tool and for insightful feedback on the cluster environment. We also thank Sanjeev Arora, Danqi Chen, Karthik Narasimhan, and other members of the Princeton Language \& Intelligence group for cooperation in our monitoring study.
\end{acks}

\FloatBarrier

\bibliographystyle{ACM-Reference-Format}
\bibliography{refs}


\begin{thebibliography}{31}


\ifx \showCODEN    \undefined \def \showCODEN     #1{\unskip}     \fi
\ifx \showISBNx    \undefined \def \showISBNx     #1{\unskip}     \fi
\ifx \showISBNxiii \undefined \def \showISBNxiii  #1{\unskip}     \fi
\ifx \showISSN     \undefined \def \showISSN      #1{\unskip}     \fi
\ifx \showLCCN     \undefined \def \showLCCN      #1{\unskip}     \fi
\ifx \shownote     \undefined \def \shownote      #1{#1}          \fi
\ifx \showarticletitle \undefined \def \showarticletitle #1{#1}   \fi
\ifx \showURL      \undefined \def \showURL       {\relax}        \fi
\providecommand\bibfield[2]{#2}
\providecommand\bibinfo[2]{#2}
\providecommand\natexlab[1]{#1}
\providecommand\showeprint[2][]{arXiv:#2}

\bibitem[Adriaens et~al\mbox{.}(2012)]%
        {adriaens_case_2012}
\bibfield{author}{\bibinfo{person}{Jacob~T. Adriaens}, \bibinfo{person}{Katherine Compton}, \bibinfo{person}{Nam~Sung Kim}, {and} \bibinfo{person}{Michael~J. Schulte}.} \bibinfo{year}{2012}\natexlab{}.
\newblock \showarticletitle{The case for {GPGPU} spatial multitasking}. In \bibinfo{booktitle}{\emph{Proceedings of the 2012 IEEE 18th International Symposium on High-Performance Computer Architecture}} \emph{(\bibinfo{series}{HPCA '12})}. \bibinfo{publisher}{IEEE Computer Society}, \bibinfo{address}{USA}, \bibinfo{pages}{1--12}.
\newblock
\showISBNx{978-1-4673-0827-4}
\href{https://doi.org/10.1109/HPCA.2012.6168946}{doi:\nolinkurl{10.1109/HPCA.2012.6168946}}


\bibitem[Chen et~al\mbox{.}(2017)]%
        {chen_effisha_2017}
\bibfield{author}{\bibinfo{person}{Guoyang Chen}, \bibinfo{person}{Yue Zhao}, \bibinfo{person}{Xipeng Shen}, {and} \bibinfo{person}{Huiyang Zhou}.} \bibinfo{year}{2017}\natexlab{}.
\newblock \showarticletitle{{EffiSha}: A Software Framework for Enabling Efficient Preemptive Scheduling of {GPU}}. In \bibinfo{booktitle}{\emph{Proceedings of the 22nd ACM SIGPLAN Symposium on Principles and Practice of Parallel Programming}} \emph{(\bibinfo{series}{PPoPP '17})}. \bibinfo{publisher}{Association for Computing Machinery}, \bibinfo{address}{New York, NY, USA}, \bibinfo{pages}{3--16}.
\newblock
\showISBNx{978-1-4503-4493-7}
\href{https://doi.org/10.1145/3018743.3018748}{doi:\nolinkurl{10.1145/3018743.3018748}}


\bibitem[Feynman(1960)]%
        {feynman_room_1960}
\bibfield{author}{\bibinfo{person}{Richard~P. Feynman}.} \bibinfo{year}{1960}\natexlab{}.
\newblock \showarticletitle{There's Plenty of Room at the Bottom}.
\newblock \bibinfo{journal}{\emph{Engineering and Science}} \bibinfo{volume}{23}, \bibinfo{number}{5} (\bibinfo{year}{1960}), \bibinfo{pages}{22--36}.
\newblock


\bibitem[Haddad(2024)]%
        {cnbc_blackwell_price}
\bibfield{author}{\bibinfo{person}{CJ Haddad}.} \bibinfo{year}{2024}\natexlab{}.
\newblock \bibinfo{title}{Nvidia {CEO} {Jensen Huang} says demand for next-generation {Blackwell AI} chip is `insane'}.
\newblock
\urldef\tempurl%
\url{https://www.cnbc.com/2024/10/03/nvidia-ceo-demand-for-blackwell-ai-chip-is-insane.html}
\showURL{%
\tempurl}


\bibitem[Hu et~al\mbox{.}(2021)]%
        {hu_characterization_2021}
\bibfield{author}{\bibinfo{person}{Qinghao Hu}, \bibinfo{person}{Peng Sun}, \bibinfo{person}{Shengen Yan}, \bibinfo{person}{Yonggang Wen}, {and} \bibinfo{person}{Tianwei Zhang}.} \bibinfo{year}{2021}\natexlab{}.
\newblock \showarticletitle{Characterization and prediction of deep learning workloads in large-scale {GPU} datacenters}. In \bibinfo{booktitle}{\emph{Proceedings of the International Conference for High Performance Computing, Networking, Storage and Analysis}} \emph{(\bibinfo{series}{SC '21})}. \bibinfo{publisher}{Association for Computing Machinery}, \bibinfo{address}{New York, NY, USA}, \bibinfo{pages}{1--15}.
\newblock
\showISBNx{978-1-4503-8442-1}
\href{https://doi.org/10.1145/3458817.3476223}{doi:\nolinkurl{10.1145/3458817.3476223}}


\bibitem[Jeon et~al\mbox{.}(2019)]%
        {jeon_analysis_2019}
\bibfield{author}{\bibinfo{person}{Myeongjae Jeon}, \bibinfo{person}{Shivaram Venkataraman}, \bibinfo{person}{Amar Phanishayee}, \bibinfo{person}{Junjie Qian}, \bibinfo{person}{Wencong Xiao}, {and} \bibinfo{person}{Fan Yang}.} \bibinfo{year}{2019}\natexlab{}.
\newblock \showarticletitle{Analysis of large-scale multi-tenant {GPU} clusters for {DNN} training workloads}. In \bibinfo{booktitle}{\emph{Proceedings of the 2019 USENIX Conference on USENIX Annual Technical Conference}} \emph{(\bibinfo{series}{USENIX ATC '19})}. \bibinfo{publisher}{USENIX Association}, \bibinfo{address}{USA}, \bibinfo{pages}{947--960}.
\newblock
\showISBNx{978-1-939133-03-8}


\bibitem[Jiang et~al\mbox{.}(2022)]%
        {jiang_moneo_2022}
\bibfield{author}{\bibinfo{person}{Yuting Jiang}, \bibinfo{person}{Yifan Xiong}, \bibinfo{person}{Lei Qu}, \bibinfo{person}{Cheng Luo}, \bibinfo{person}{Chen Tian}, \bibinfo{person}{Peng Cheng}, {and} \bibinfo{person}{Yongqiang Xiong}.} \bibinfo{year}{2022}\natexlab{}.
\newblock \showarticletitle{Moneo: Monitoring Fine-grained Metrics Nonintrusively in {AI} Infrastructure}.
\newblock \bibinfo{journal}{\emph{ACM SIGOPS Operating Systems Review}} \bibinfo{volume}{56}, \bibinfo{number}{1} (\bibinfo{date}{June} \bibinfo{year}{2022}), \bibinfo{pages}{18--25}.
\newblock
\showISSN{0163-5980}
\href{https://doi.org/10.1145/3544497.3544501}{doi:\nolinkurl{10.1145/3544497.3544501}}


\bibitem[Leiserson et~al\mbox{.}(2020)]%
        {leiserson_room_2020}
\bibfield{author}{\bibinfo{person}{Charles~E. Leiserson}, \bibinfo{person}{Neil~C. Thompson}, \bibinfo{person}{Joel~S. Emer}, \bibinfo{person}{Bradley~C. Kuszmaul}, \bibinfo{person}{Butler~W. Lampson}, \bibinfo{person}{Daniel Sanchez}, {and} \bibinfo{person}{Tao~B. Schardl}.} \bibinfo{year}{2020}\natexlab{}.
\newblock \showarticletitle{There's Plenty of Room at the Top: What Will Drive Computer Performance after {Moore's} Law?}
\newblock \bibinfo{journal}{\emph{Science}} \bibinfo{volume}{368}, \bibinfo{number}{6495} (\bibinfo{year}{2020}), \bibinfo{pages}{eaam9744}.
\newblock
\href{https://doi.org/10.1126/science.aam9744}{doi:\nolinkurl{10.1126/science.aam9744}}


\bibitem[Liang et~al\mbox{.}(2015)]%
        {yun_liang_efficient_2015}
\bibfield{author}{\bibinfo{person}{Yun Liang}, \bibinfo{person}{Huynh~Phung Huynh}, \bibinfo{person}{Kyle Rupnow}, \bibinfo{person}{Rick Siow~Mong Goh}, {and} \bibinfo{person}{Deming Chen}.} \bibinfo{year}{2015}\natexlab{}.
\newblock \showarticletitle{Efficient {GPU} Spatial-Temporal Multitasking}.
\newblock \bibinfo{journal}{\emph{IEEE Transactions on Parallel and Distributed Systems}} \bibinfo{volume}{26}, \bibinfo{number}{3} (\bibinfo{year}{2015}), \bibinfo{pages}{748--760}.
\newblock
\showISSN{1045-9219}
\href{https://doi.org/10.1109/TPDS.2014.2313342}{doi:\nolinkurl{10.1109/TPDS.2014.2313342}}


\bibitem[Mahajan et~al\mbox{.}(2020)]%
        {mahajan_themis_2020}
\bibfield{author}{\bibinfo{person}{Kshiteej Mahajan}, \bibinfo{person}{Arjun Balasubramanian}, \bibinfo{person}{Arjun Singhvi}, \bibinfo{person}{Shivaram Venkataraman}, \bibinfo{person}{Aditya Akella}, \bibinfo{person}{Amar Phanishayee}, {and} \bibinfo{person}{Shuchi Chawla}.} \bibinfo{year}{2020}\natexlab{}.
\newblock \showarticletitle{Themis: Fair and Efficient {GPU} Cluster Scheduling}. In \bibinfo{booktitle}{\emph{17th USENIX Symposium on Networked Systems Design and Implementation (NSDI 20)}}. \bibinfo{publisher}{USENIX Association}, \bibinfo{pages}{289--304}.
\newblock
\showISBNx{978-1-939133-13-7}
\urldef\tempurl%
\url{https://www.usenix.org/conference/nsdi20/presentation/mahajan}
\showURL{%
\tempurl}


\bibitem[Massie et~al\mbox{.}(2004)]%
        {massie_ganglia_2004}
\bibfield{author}{\bibinfo{person}{Matthew~L. Massie}, \bibinfo{person}{Brent~N. Chun}, {and} \bibinfo{person}{David~E. Culler}.} \bibinfo{year}{2004}\natexlab{}.
\newblock \showarticletitle{The Ganglia distributed monitoring system: design, implementation, and experience}.
\newblock \bibinfo{journal}{\emph{Parallel Comput.}} \bibinfo{volume}{30}, \bibinfo{number}{7} (\bibinfo{date}{July} \bibinfo{year}{2004}), \bibinfo{pages}{817--840}.
\newblock
\showISSN{0167-8191}
\href{https://doi.org/10.1016/j.parco.2004.04.001}{doi:\nolinkurl{10.1016/j.parco.2004.04.001}}


\bibitem[{Nvidia}(2007)]%
        {nvidia_nvml}
\bibfield{author}{\bibinfo{person}{{Nvidia}}.} \bibinfo{year}{2007}\natexlab{}.
\newblock \bibinfo{title}{{NVIDIA} Management Library ({NVML})}.
\newblock
\urldef\tempurl%
\url{https://developer.nvidia.com/management-library-nvml}
\showURL{%
\tempurl}


\bibitem[{Nvidia}(2017)]%
        {nvidia_uvm}
\bibfield{author}{\bibinfo{person}{{Nvidia}}.} \bibinfo{year}{2017}\natexlab{}.
\newblock \bibinfo{title}{Unified Memory for {CUDA} Beginners}.
\newblock
\urldef\tempurl%
\url{https://developer.nvidia.com/blog/unified-memory-cuda-beginners/}
\showURL{%
\tempurl}


\bibitem[{Nvidia}(2020a)]%
        {nvidia_mig}
\bibfield{author}{\bibinfo{person}{{Nvidia}}.} \bibinfo{year}{2020}\natexlab{a}.
\newblock \bibinfo{title}{Multi-Instance {GPU} User Guide}.
\newblock
\urldef\tempurl%
\url{https://docs.nvidia.com/datacenter/tesla/pdf/NVIDIA_MIG_User_Guide.pdf}
\showURL{%
\tempurl}


\bibitem[{Nvidia}(2020b)]%
        {nvidia_mps}
\bibfield{author}{\bibinfo{person}{{Nvidia}}.} \bibinfo{year}{2020}\natexlab{b}.
\newblock \bibinfo{title}{Multi-Process Service User Guide}.
\newblock
\urldef\tempurl%
\url{https://docs.nvidia.com/deploy/mps/index.html}
\showURL{%
\tempurl}


\bibitem[{Nvidia}(2020c)]%
        {nvidia_dcgm}
\bibfield{author}{\bibinfo{person}{{Nvidia}}.} \bibinfo{year}{2020}\natexlab{c}.
\newblock \bibinfo{title}{{NVIDIA DCGM} Documentation}.
\newblock
\urldef\tempurl%
\url{https://docs.nvidia.com/datacenter/dcgm/latest/index.html}
\showURL{%
\tempurl}


\bibitem[{Nvidia}(2024a)]%
        {nvidia_cuda_programming_12_4_1}
\bibfield{author}{\bibinfo{person}{{Nvidia}}.} \bibinfo{year}{2024}\natexlab{a}.
\newblock \bibinfo{title}{{CUDA} C++ Programming Guide (Version 12.4.1)}.
\newblock
\urldef\tempurl%
\url{https://docs.nvidia.com/cuda/archive/12.4.1/cuda-c-programming-guide/index.html}
\showURL{%
\tempurl}


\bibitem[{Nvidia}(2024b)]%
        {nvidia_cuda_runtime_12_4_1}
\bibfield{author}{\bibinfo{person}{{Nvidia}}.} \bibinfo{year}{2024}\natexlab{b}.
\newblock \bibinfo{title}{{CUDA} Runtime {API}: Memory Management (Version 12.4.1)}.
\newblock
\urldef\tempurl%
\url{https://docs.nvidia.com/cuda/archive/12.4.1/cuda-runtime-api/group__CUDART__MEMORY.html}
\showURL{%
\tempurl}


\bibitem[Park et~al\mbox{.}(2015)]%
        {park_chimera_2015}
\bibfield{author}{\bibinfo{person}{Jason Jong~Kyu Park}, \bibinfo{person}{Yongjun Park}, {and} \bibinfo{person}{Scott Mahlke}.} \bibinfo{year}{2015}\natexlab{}.
\newblock \showarticletitle{Chimera: Collaborative Preemption for Multitasking on a Shared {GPU}}. In \bibinfo{booktitle}{\emph{Proceedings of the Twentieth International Conference on Architectural Support for Programming Languages and Operating Systems (ASPLOS '15)}}. \bibinfo{publisher}{Association for Computing Machinery}, \bibinfo{address}{New York, NY, USA}, \bibinfo{pages}{593--606}.
\newblock
\href{https://doi.org/10.1145/2775054.2694346}{doi:\nolinkurl{10.1145/2775054.2694346}}


\bibitem[Rabenstein and Volz(2015)]%
        {prometheus}
\bibfield{author}{\bibinfo{person}{Bj{\"o}rn Rabenstein} {and} \bibinfo{person}{Julius Volz}.} \bibinfo{year}{2015}\natexlab{}.
\newblock \showarticletitle{Prometheus: A Next-Generation Monitoring System (Talk)}. In \bibinfo{booktitle}{\emph{SREcon15 Europe}}. \bibinfo{publisher}{USENIX Association}, \bibinfo{address}{Dublin}.
\newblock


\bibitem[Samsi et~al\mbox{.}(2021)]%
        {samsi_supercloud_2021}
\bibfield{author}{\bibinfo{person}{Siddharth Samsi}, \bibinfo{person}{Matthew~L. Weiss}, \bibinfo{person}{David Bestor}, \bibinfo{person}{Baolin Li}, \bibinfo{person}{Michael Jones}, \bibinfo{person}{Albert Reuther}, \bibinfo{person}{Daniel Edelman}, \bibinfo{person}{William Arcand}, \bibinfo{person}{Chansup Byun}, \bibinfo{person}{John Holodnack}, \bibinfo{person}{Matthew Hubbell}, \bibinfo{person}{Jeremy Kepner}, \bibinfo{person}{Anna Klein}, \bibinfo{person}{Joseph McDonald}, \bibinfo{person}{Adam Michaleas}, \bibinfo{person}{Peter Michaleas}, \bibinfo{person}{Lauren Milechin}, \bibinfo{person}{Julia Mullen}, \bibinfo{person}{Charles Yee}, \bibinfo{person}{Benjamin Price}, \bibinfo{person}{Andrew Prout}, \bibinfo{person}{Antonio Rosa}, \bibinfo{person}{Allan Vanterpool}, \bibinfo{person}{Lindsey McEvoy}, \bibinfo{person}{Anson Cheng}, \bibinfo{person}{Devesh Tiwari}, {and} \bibinfo{person}{Vijay Gadepally}.} \bibinfo{year}{2021}\natexlab{}.
\newblock \showarticletitle{The {MIT} Supercloud Dataset}. In \bibinfo{booktitle}{\emph{2021 IEEE High Performance Extreme Computing Conference (HPEC)}}. \bibinfo{pages}{1--8}.
\newblock
\href{https://doi.org/10.1109/HPEC49654.2021.9622850}{doi:\nolinkurl{10.1109/HPEC49654.2021.9622850}}


\bibitem[Schmuck and Haskin(2002)]%
        {gpfs}
\bibfield{author}{\bibinfo{person}{Frank Schmuck} {and} \bibinfo{person}{Roger Haskin}.} \bibinfo{year}{2002}\natexlab{}.
\newblock \showarticletitle{{GPFS}, a Shared-Disk File System for Large Computing Clusters}. In \bibinfo{booktitle}{\emph{Proceedings of the FAST 2002 Conference on File and Storage Technologies}}. \bibinfo{address}{Monterey, California}.
\newblock


\bibitem[Sedighi et~al\mbox{.}(2024)]%
        {sedighi_efficient_2024}
\bibfield{author}{\bibinfo{person}{Hoda Sedighi}, \bibinfo{person}{Daniel Gehberger}, \bibinfo{person}{Amin Ebrahimzadeh}, \bibinfo{person}{Fetahi Wuhib}, {and} \bibinfo{person}{Roch~H. Glitho}.} \bibinfo{year}{2024}\natexlab{}.
\newblock \showarticletitle{Efficient Dynamic Resource Management for Spatial Multitasking {GPUs}}.
\newblock \bibinfo{journal}{\emph{IEEE Transactions on Cloud Computing}} (\bibinfo{year}{2024}), \bibinfo{pages}{1--18}.
\newblock
\showISSN{2168-7161}
\href{https://doi.org/10.1109/TCC.2024.3511548}{doi:\nolinkurl{10.1109/TCC.2024.3511548}}


\bibitem[Strati et~al\mbox{.}(2024)]%
        {strati_orion_2024}
\bibfield{author}{\bibinfo{person}{Foteini Strati}, \bibinfo{person}{Xianzhe Ma}, {and} \bibinfo{person}{Ana Klimovic}.} \bibinfo{year}{2024}\natexlab{}.
\newblock \showarticletitle{Orion: Interference-aware, Fine-grained {GPU} Sharing for {ML} Applications}. In \bibinfo{booktitle}{\emph{Proceedings of the Nineteenth European Conference on Computer Systems}} \emph{(\bibinfo{series}{EuroSys '24})}. \bibinfo{publisher}{Association for Computing Machinery}, \bibinfo{address}{New York, NY, USA}, \bibinfo{pages}{1075--1092}.
\newblock
\showISBNx{979-8-4007-0437-6}
\href{https://doi.org/10.1145/3627703.3629578}{doi:\nolinkurl{10.1145/3627703.3629578}}


\bibitem[Tanasic et~al\mbox{.}(2014)]%
        {tanasic_enabling_2014}
\bibfield{author}{\bibinfo{person}{Ivan Tanasic}, \bibinfo{person}{Isaac Gelado}, \bibinfo{person}{Javier Cabezas}, \bibinfo{person}{Alex Ramirez}, \bibinfo{person}{Nacho Navarro}, {and} \bibinfo{person}{Mateo Valero}.} \bibinfo{year}{2014}\natexlab{}.
\newblock \showarticletitle{Enabling preemptive multiprogramming on {GPUs}}. In \bibinfo{booktitle}{\emph{Proceeding of the 41st Annual International Symposium on Computer Architecture (ISCA '14)}}. \bibinfo{publisher}{IEEE Press}, \bibinfo{pages}{193--204}.
\newblock
\href{https://doi.org/10.1145/2678373.2665702}{doi:\nolinkurl{10.1145/2678373.2665702}}


\bibitem[Wang et~al\mbox{.}(2016)]%
        {wang_simultaneous_2016}
\bibfield{author}{\bibinfo{person}{Zhenning Wang}, \bibinfo{person}{Jun Yang}, \bibinfo{person}{Rami Melhem}, \bibinfo{person}{Bruce Childers}, \bibinfo{person}{Youtao Zhang}, {and} \bibinfo{person}{Minyi Guo}.} \bibinfo{year}{2016}\natexlab{}.
\newblock \showarticletitle{Simultaneous Multikernel {GPU}: Multi-tasking throughput processors via fine-grained sharing}. In \bibinfo{booktitle}{\emph{2016 IEEE International Symposium on High Performance Computer Architecture (HPCA)}}. \bibinfo{publisher}{IEEE}, \bibinfo{pages}{358--369}.
\newblock
\href{https://doi.org/10.1109/HPCA.2016.7446078}{doi:\nolinkurl{10.1109/HPCA.2016.7446078}}


\bibitem[Weakley et~al\mbox{.}(2025)]%
        {weakley_monitoring_2025}
\bibfield{author}{\bibinfo{person}{Le~Mai Weakley}, \bibinfo{person}{Scott Michael}, \bibinfo{person}{Laura Huber}, \bibinfo{person}{Abhinav Thota}, \bibinfo{person}{Ben Fulton}, {and} \bibinfo{person}{Matthew Kusz}.} \bibinfo{year}{2025}\natexlab{}.
\newblock \showarticletitle{Monitoring and Characterizing {GPU} Usage}.
\newblock \bibinfo{journal}{\emph{Concurrency and Computation: Practice and Experience}} \bibinfo{volume}{37}, \bibinfo{number}{3} (\bibinfo{year}{2025}), \bibinfo{pages}{e8341}.
\newblock
\showISSN{1532-0634}
\href{https://doi.org/10.1002/cpe.8341}{doi:\nolinkurl{10.1002/cpe.8341}}


\bibitem[Weng et~al\mbox{.}(2022)]%
        {weng_mlaas_2022}
\bibfield{author}{\bibinfo{person}{Qizhen Weng}, \bibinfo{person}{Wencong Xiao}, \bibinfo{person}{Yinghao Yu}, \bibinfo{person}{Wei Wang}, \bibinfo{person}{Cheng Wang}, \bibinfo{person}{Jian He}, \bibinfo{person}{Yong Li}, \bibinfo{person}{Liping Zhang}, \bibinfo{person}{Wei Lin}, {and} \bibinfo{person}{Yu Ding}.} \bibinfo{year}{2022}\natexlab{}.
\newblock \showarticletitle{{MLaaS} in the Wild: Workload Analysis and Scheduling in Large-Scale Heterogeneous {GPU} Clusters}. In \bibinfo{booktitle}{\emph{19th USENIX Symposium on Networked Systems Design and Implementation (NSDI 22)}}. \bibinfo{publisher}{USENIX Association}, \bibinfo{pages}{945--960}.
\newblock
\showISBNx{978-1-939133-27-4}
\urldef\tempurl%
\url{https://www.usenix.org/conference/nsdi22/presentation/weng}
\showURL{%
\tempurl}


\bibitem[Wu et~al\mbox{.}(2024)]%
        {wu2024falconpinpointingmitigatingstragglers}
\bibfield{author}{\bibinfo{person}{Tianyuan Wu}, \bibinfo{person}{Wei Wang}, \bibinfo{person}{Yinghao Yu}, \bibinfo{person}{Siran Yang}, \bibinfo{person}{Wenchao Wu}, \bibinfo{person}{Qinkai Duan}, \bibinfo{person}{Guodong Yang}, \bibinfo{person}{Jiamang Wang}, \bibinfo{person}{Lin Qu}, {and} \bibinfo{person}{Liping Zhang}.} \bibinfo{year}{2024}\natexlab{}.
\newblock \bibinfo{title}{{FALCON}: Pinpointing and Mitigating Stragglers for Large-Scale Hybrid-Parallel Training}.
\newblock
\showeprint[arxiv]{2410.12588}~[cs.DC]
\urldef\tempurl%
\url{https://arxiv.org/abs/2410.12588}
\showURL{%
\tempurl}


\bibitem[Xiao et~al\mbox{.}(2018)]%
        {xiao_gandiva_2018}
\bibfield{author}{\bibinfo{person}{Wencong Xiao}, \bibinfo{person}{Romil Bhardwaj}, \bibinfo{person}{Ramachandran Ramjee}, \bibinfo{person}{Muthian Sivathanu}, \bibinfo{person}{Nipun Kwatra}, \bibinfo{person}{Zhenhua Han}, \bibinfo{person}{Pratyush Patel}, \bibinfo{person}{Xuan Peng}, \bibinfo{person}{Hanyu Zhao}, \bibinfo{person}{Quanlu Zhang}, \bibinfo{person}{Fan Yang}, {and} \bibinfo{person}{Lidong Zhou}.} \bibinfo{year}{2018}\natexlab{}.
\newblock \showarticletitle{Gandiva: Introspective Cluster Scheduling for Deep Learning}. In \bibinfo{booktitle}{\emph{13th USENIX Symposium on Operating Systems Design and Implementation (OSDI 18)}}. \bibinfo{publisher}{USENIX Association}, \bibinfo{pages}{595--610}.
\newblock
\showISBNx{978-1-939133-08-3}
\urldef\tempurl%
\url{https://www.usenix.org/conference/osdi18/presentation/xiao}
\showURL{%
\tempurl}


\bibitem[Xiao et~al\mbox{.}(2020)]%
        {xiao_antman_2020}
\bibfield{author}{\bibinfo{person}{Wencong Xiao}, \bibinfo{person}{Shiru Ren}, \bibinfo{person}{Yong Li}, \bibinfo{person}{Yang Zhang}, \bibinfo{person}{Pengyang Hou}, \bibinfo{person}{Zhi Li}, \bibinfo{person}{Yihui Feng}, \bibinfo{person}{Wei Lin}, {and} \bibinfo{person}{Yangqing Jia}.} \bibinfo{year}{2020}\natexlab{}.
\newblock \showarticletitle{{AntMan}: dynamic scaling on {GPU} clusters for deep learning}. In \bibinfo{booktitle}{\emph{Proceedings of the 14th USENIX Conference on Operating Systems Design and Implementation}} \emph{(\bibinfo{series}{OSDI '20})}. \bibinfo{publisher}{USENIX Association}, \bibinfo{address}{USA}, \bibinfo{pages}{533--548}.
\newblock
\showISBNx{978-1-939133-19-9}


\end{thebibliography}

\end{document}